\documentclass[12pt]{amsart}

\usepackage[lite]{amsrefs}

\usepackage{amssymb}

\usepackage[all,cmtip]{xy}

\usepackage[ruled,vlined,linesnumbered]{algorithm2e}
\let\oldnl\nl% Store \nl in \oldnl
\newcommand{\nonl}{\renewcommand{\nl}{\let\nl\oldnl}}
\usepackage[noend]{algpseudocode}
\usepackage{graphicx}
\usepackage[framemethod=TikZ]{mdframed}
\usepackage{lipsum}
\usepackage{hyperref}
\definecolor{linkcolour}{rgb}{0,0.2,0.6}
\hypersetup{colorlinks,breaklinks,urlcolor=linkcolour, linkcolor=linkcolour}
\usepackage{url}
\usepackage{amsmath}   
\usepackage[english]{babel}
\usepackage{subfig}
\usepackage[ruled,vlined,linesnumbered]{algorithm2e}
\let\oldnl\nl% Store \nl in \oldnl
\usepackage[noend]{algpseudocode}
\usepackage{amssymb,bm}
\usepackage{geometry}
\usepackage{framed}
\usepackage{latexsym}
\usepackage{setspace}
\usepackage{cite}
\usepackage{fullpage}
\usepackage[tableposition=above]{caption}
\RequirePackage[english=usenglishmax]{hyphsubst}
\numberwithin{equation}{section}

\theoremstyle{plain} %% This is the default, anyway

\theoremstyle{definition}

\theoremstyle{remark}

\begin{document}

\begin{center}

\title{Learning low dimensional word based linear classifiers using Data Shared Adaptive Bootstrap Aggregated Lasso with application to IMDb data}

\end{center}

\author{Ashutosh K. Maurya$^{[1,2]}$\\Applied Statistics Unit,\\Indian Statistical Institute, Kolkata}

 \curraddr{}

%%% To have the current date inserted, use \date{\today}:
\date{\today}
\footnotetext[1]{ \* Author: Postal Address: Applied Statistics Unit, Indian Statistical Institute, Kolkata, India, Tel.: +91 9123305986, Email: ashutoshmaurya2010@gmail.com$^{[1]}$}
\footnotetext[2]{\* Project partially supported by Statistical Trainee program, Applied Statistics Unit, ISI, Kolkata and the R. C. Bose Centre for Cryptology and Security, ISI, Kolkata.}
%%% To include an abstract, uncomment the following two lines and type
%%% the abstract in between them:
 \begin{abstract}

 In this article we propose a new supervised ensemble learning method called Data Shared Adaptive Bootstrap Aggregated (AdaBag) Lasso for capturing low dimensional useful features for word based sentiment analysis and mining problems (\cite{pang}). The literature on ensemble methods is very rich in both statistics and machine learning (\cite{bagging},\cite{hastie},\cite{knox}).
 
 %and perform dimension reduction of a large dataset which may or may not be grouped.
 The algorithm is a substantial upgrade of  the Data Shared Lasso uplift algorithm (\cite{gross}). The most significant conceptual addition to the existing literature lies in the final selection of bag of predictors through a special bootstrap aggregation scheme.
 
 AdaBag is intuitive, modular and flexible in nature. All computations have been done using R package (  Matrix, glmnet \cite{friedman}, wordcloud,  ggplot2, MASS \textit{etc.}) in PC (Windows and Ubuntu). The primary source of reference is 
 \href{https://www.rdocumentation.org/}{R Documentation}.
 
  We apply the algorithm to one simulated  data and perform dimension reduction in grouped IMDb data (drama, comedy and horror, \cite{mass}) to extract reduced set of word features for predicting sentiment ratings of movie reviews demonstrating different aspects. We also compare the performance of the present method with the classical Principal Components with associated Linear Discrimination (PCA-LD) as baseline (\cite{pcalda},\cite{kuhn}).
  
  There are few limitations in the algorithm. Firstly, the algorithm workflow does not incorporate online sequential data acquisition and it does not use sentence based models which are common in ANN algorithms (\cite{topic}). Our results produce slightly higher error rate compare to the reported state-of-the-art (\cite{miyato}) as a consequence.

 $\bf Key~ Words:$ 
 
 Supervised Dimension Reduction, Adaptive Lasso, Bootstrap Aggregation, Principal Components, Linear Discrimination, Data Shared Lasso.
 \end{abstract}

\maketitle

\section{Introduction}

Mining important features and predictors from big data in supervised learning is a challenging task. Basically it requires two key steps: dimension reduction and computation of well performing predictors. Both the steps are complex and computationally involved.

There are a number of algorithms for feature selection. A host of useful methodology exists,
 \cite{chen} (data enriched regression) \cite{hoerl}(ridge regression), subset selection (\cite{sara}), \cite{lars} (least angle regression),
  \cite{donoho}) (wavelet De-noising) and \cite{tibshirani} (least absolute shrinkage and selection operator) to name a few. We skip referencing to a huge body of methodology available in machine learning (ML), artificial neural networks (ANN) and natural language processing (NLP) here.

The IMDb problem (\cite{mass}) can be viewed as an application of sentiment analysis where the goal is to extract a smaller set of features from very large \textit{bag-of-words} that have better classification of movie ratings. This template example can serve as a prototype of a large number of similar problems of sentiment analysis in different areas (\cite{pang},\cite{kuhn},\cite{hastie},\cite{knox}). The IMDb dataset\cite{mass} consist of 50k movie reviews on a scale of allowing no more than 30 reviews per movie where response variable is integer rating in the scale $1$ to $10$ where ratings ($>6$) are taken to be high positive ratings and ratings ($<5$) are taken to be high negative ratings to make it a binary class classification problem and predictors are those words which appear in atleast ($5$) reviews. 

In \cite{gross}, a method of data shared lasso (DSL) has been developed and analyzed using the mean squared error.
In AdaBag procedure, we use classification error rate while leaving open the option of data sharing (in case of grouped data). The choice of adaptive tuning in the algorithm is left to the user. In short AdaBag can be described as follows.
\begin{enumerate}
    \item[$(i)$] \textit{Initialization}. A collection of low dimensional word models are produced from the user supplied adaptive Lasso formulation from core training set. 
    \item[$(ii)$] \textit{Choice of regularization parameter}. From the above collection of submodels validate their performance (after tuning within-model parameters using OLS) in the validation part of the training set to choose the optimum value of the regularization parameter (\cite{bickel}) and corresponding word model.
    \item[$(iii)$] \textit{Bagging}. Bootstrap $(i)$ $\&$ $(ii)$ repeated $B$ times and aggregate this low dimensional models (corresponding to the optimum value of the regularization parameter in $(ii)$ in each run) to create a \textit{bagging frequency distribution}(\textit{bfd}) across the \textit{bow} by adding the incidences of words to obtain a word frequency (\cite{efron},\cite{efron1},\cite{bagging},\cite{hall}). 
    
    \item[$(iv)$] \textit{Sorting word relevance}. The higher the frequency of a word in the \textit{bfd} the more relevant it is. 
    Finally class of models are defined by subset of words in \textit{bfd} having relevance more than different levels of cutoffs.
    Sort the \textit{bfd} according to relevance of words. The AdaBag chosen models is a sequence of nested models determined by words in \textit{bfd} having relevance more than different levels of cutoff.
\end{enumerate}

In IMDb data application where movies are rated through text based reviews the bow size is $p = 27743$. There are  $n=50k$ movie reviews divided into three categories, drama comedy and horror. Thus for IMDb data for each group $n<p$, however $n=O(p)$. For computational purposes the data is further organized into three subsets namely, \textit{core}, \textit{validation} and \textit{test}. The core $(50\%)$ and validation $(25\%)$ constitute the training set and the remaining $25\%$  are left for testing purposes (according to positive and negative movie ratings). We use \textit{polarity} of words in \textit{bow} as auxiliary information to construct the adaptive Lasso regularization (\cite{adlasso}). Computation of polarity of words for IMDb corpus has been extensively studied by \cite{potts}. In this work \textit{word weights are chosen inversely proportional to polarity}. The present article is an upgrade of previous idea which was based on mean squared error(\cite{maurya}).

\textit{Why didn't we use the traditional cross validation?} We found the cross validation module \textit{cv.glmnet}, for optimizing the regularization parameter to be unsuitable for our purpose when we compute classification error rate. The codes and options are not flexible enough to incorporate different user supplied performance criteria. Because of this reason we introduce a separate validation set to choose optimal regularization parameter. The idea of a validation set turns out to be very useful as it acts like a pseudo test sample within the training set. As a consequence it becomes much easier to optimize the regularization parameter under different types of performance criteria such as precision and recall. We found that computation time actually reduces if we use the validation set approach compare to cross validation.

Note that for high dimensional problems like IMDb it is difficult to rationalize the existence of a true model. Theoretical results (for example, Theorem 6.1 in \cite{sara}) assume the existence of a sparse active set (\textit{truth}). Moreover an oracle type inequality for adaptive Lasso can be found in Theorem 2 of \cite{zou1}. 

The basic novelty of our approach lies in adapting polarity information of words and devising the idea of \textit{bfd} to construct a nested sequence of word sub models which should contain models close to Oracle bound (\cite{oracle}) in sparse high dimensional classification problems.

Following the notations of Section \ref{prel} for principal component analysis of IMDb data we consider a training set which is the union of core and the validation sets ($75\%$ of the data). Sparse PCA methodology is discussed in \cite{sparsepca}. The reason is there is no regularization parameter to be minimized in this algorithm. The design matrix $X$ consists of rows $\mathbf{x}_i^T$ for $i \in \mathcal{T}_r$. Using \textit{princomp} in \textbf{R} library with default options (centered and scaled) to obtain the eigenvalues ($\omega_1\geq\omega_2\geq\ldots$) and eigenvectors ($\mathbf{u}_1,\mathbf{u}_2,\cdots$) of the singular value decomposition of $X$. The features computed for PCA transformation are given by $\mathbf{u}_1^{T}\mathbf{x},\,\mathbf{u}_2^{T}\mathbf{x},\,\ldots$ . There are few problems with principal component approach. Although $\omega_1>\omega_2>\cdots$ is an efficient variance decomposition of the data it is with respect to a $\ell_2$ metric. Here we are interested in misclassification which is primarily linked with $\ell_1$ metric. As a consequence the features $\mathbf{u}_i^{T}\mathbf{x}$ provides certain linear combinations of word indicator variables in a review. Such linear combinations assigns non zero weights almost across the \textit{bow} and thus hard to interpret. For this reason we also consider a variant of the usual PCA where $\mathbf{u}_i$'s are ordered according to the entropy of an artificial probability ($u_{i1}^2,u_{i2}^2,\ldots$) on \textit{bow}.

More serious problem however is related to thresholding. For dimension reduction (and computation of classifier using linear discrimination formula in the next step) we need to fewer number of features from the rotation matrix. However that has to be traded off with proportion of variance explained using $\omega_1>\omega_2>\cdots$. There is no universally accepted trade-off formula as in the case of mean squared error in nonparametric function estimation. From what we found in the case of IMDb data such a trade-off may heavily depend on the specifics of the sample at hand rather than the universe of the possible samples (average properties of the corpus). Based on our experience with PCA with IMDb data we fix the percent of variance explained at $30\%$. This amounts to about top $150-200$ features.

There are other approaches in ML and ANN literature where one delves into deeper learning of using \textit{sentence model}  in NLP. A summary of such results can be found in (\cite{topic},\cite{miyato}). There are several other model selection techniques like subset regression, AIC, BIC, etc. which we will attend to in greater detail in Section \ref{prel}.

The article is presented as follows: In Section $2$ we present basic theoretical preliminaries about variable selection of high dimensional data. In this Section we discuss Lasso and other related regularization methods used to motivate AdaBag. In Section $3$ we illustrate our proposed methodology for the analyzing grouped data and also consider a simulated data. In Section $5$ we talk about the preparation of the dataset and output of data analysis is reported and discussed and finally concluding remark is given in Section $6$.

\section{Preliminaries}
\label{prel}

Suppose we have $n$ observations from a grouped database with $G$ groups of the form  $( y_i,\mathbf{x}_i^T ,g_i)$ where $\mathbf{x}_i \in \mathbb{R}^p$, $g_i \in \mathbb{G}= \{1,2,\cdots,G\}$. The response variable $y_i$'s can be categorical, ordinal or continuous. To fix ideas we assume a multiple linear regression setup. However the data sharing scheme (as in \cite{gross}) can be  generalized linear models such as logistic regression. Further we transform the response variables assuming they are either ordinal or continuous in two classes namely, $I(y\leq a)$ and $I(y\geq b)$ for two preset levels $a<b$. In a prediction scenario where $y$ is unobserved we introduce a class indicator $\theta$ as

\begin{align}
    \theta=\Big\{^{1\quad if\ y\geq b}_{0\quad if\ y\leq a}.
    \label{class1}
\end{align}

In general in supervised learning problems a sample is split into two parts: training ($\mathcal{T}_r$) and test subsets($\mathcal{T}_s$). In training samples values $y_i$'s of the response variable are known. For the test set the data is of the form $(\theta_i,\mathbf{x}_i,g_i)$ for $i \in \mathcal{T}_s$.
Learning algorithms are developed using samples in $T_r$ in the form of a prediction rule $\phi(\mathcal{T}_r,\cdot): \mathbb{R}^p \times \mathbb{G} \longrightarrow \{0,1\}$. When we input a value $\mathbf{x}$, of the explanatory variable a particular class $\phi(\mathcal{T}_r,\mathbf{x},g) \in \{0,1\}$ is predicted. Thus, training subset helps us learn about the unknown $\theta_i$ using  $\phi(\mathcal{T}_r,\mathbf{x}_i,g_i)$ for $i \in \mathcal{T}_s$.

Prediction in $\mathcal{T}_s$ using $\mathcal{T}_r$ generates prediction error which is measured by $|\theta_i - \phi(\mathcal{T}_r,\mathbf{x}_i,g_i)|$ for $i\in \mathcal{T}_s$. The average error 
$$\frac{1}{k}\, \sum_{i=1}^{k} \, |\theta_i - \phi(\mathcal{T}_r,\mathbf{x}_i,g_i)|, $$ 
where $k=|\mathcal{T}_s|$, is an estimate of the theoretical \textit{misclassification error}. When $\phi$ does not depend on $g$ we call it a \textit{shrinkage predictor} (\cite{tibshirani}). In a similar vein if one can find a vector $\mathbf{b}$ such that $\phi(\mathcal{T}_r,\mathbf{x},g)= \phi(\mathcal{T}_r,\mathbf{x}^T \mathbf{b},g)$ then $\phi$ is called a \textit{linear predictor}.

In this study we consider a supervised learning problem, where the dataset is split into three parts: training $\mathcal{T}_r$, validation $\mathcal{V}_d$ and test $\mathcal{T}_s$ subsets. The explanatory variables are $p$ dimensional with ($p>>n$). Moreover here we focus on the situation where the design matrix $X= \left(\,(x_{ij})\,\right) $ is binary and sparse. The response variable have been converted into binary variable for the validation of the model and to check the predication accuracy of model in test set.

 \noindent \textit{Construction of the classifier.}
 Consider the usual $0-1$ loss function. Then for an action $\delta \in \{0,1\}$
\begin{align*}
E[L(\theta,\delta)| (\mathbf{x},g)]&=E[L(\mathcal{I}(y\geq b)-\mathcal{I}(y\leq a),\delta) | (\mathbf{x},g)]  \nonumber \\
       &=L(1,\delta). P(y\geq b| (\mathbf{x},g))+   L(0,\delta). P(y\leq a| (\mathbf{x},g))
       %& \phantom{==}+L(0,A)P(a<y<b) \nonumber
\end{align*}

From here it follows that the optimum classifier is given by (ignoring any prior knowledge regarding true probabilities of observations belonging to different classes)
\begin{align}
    \hat{y}&= \phi(\mathcal{T}_r,\mathbf{x},g)= \Big\{^{1 \ if \  P(y\geq b| (\mathbf{x},g)) >= P(y\leq a| (\mathbf{x},g))}_{0 \ if \ P(y\geq b| (\mathbf{x},g)) < {P(y\leq a| (\mathbf{x},g)) .}}
    \label{mis_rule}
\end{align}
The conditional probabilities are user specified. In this article we assume a Gaussian model for conditional distribution in both classes. Thus.\\ $ P(y\geq b| (\mathbf{x},g)) =1- \Phi\left(\frac{b-x'\hat{\mathbf{b}}_g}{\hat{\sigma}_g}\right)$  and $ {P(y\leq a| (\mathbf{x},g))}= \Phi\left(\frac{a-x'\hat{\mathbf{b}}_g}{\hat{\sigma}_g}\right)$ where where $\Phi$ is standard normal CDF and ($\hat{b}_g,\hat{\sigma}_g$), $g=1,2,\ldots, G$ are separate OLS estimates within group. Traditionally this is probit classifier. Another popular choice for the conditional model is logistic distribution in two class problem. For further information about classical generalized linear model and it's nonparametric version we refer to (\cite{agresti},\cite{adaboost}).

In this article we consider the following linear regression model to calculate the predicted quantities under Gaussian error.
\begin{equation}
    \label{dslopt}
     y_{i} =\mu_{g_i}+ \mathbf{x}_{i}^{T}(\boldsymbol{\beta} + \mathbf{\Delta}_{g_{i}}) + \epsilon_{i},
\end{equation}

 where $\mu_{g_i}$ is common intercept within each group $g$,$g\in \mathbb{G}$and $\epsilon_{i}$'s are iid with mean zero and variance $\sigma_{g_i}^{2}$. Note that (\ref{dslopt}) is simply a separate regression model in different groups. If the parameter $\mathbf{\Delta}$ is absent then it is a shrinkage model when $\mu_1=\mu_2=\ldots=\mu_G$, 
 which shrinks data towards a common regression model. The idea of shrinkage and its benefits was introduced in the classic paper of \cite{stein}.

It is well known that ordinary least square (OLS) estimation procedure without appropriate regularization performs poorly in both dimension reduction and prediction when ($p>>n$). To overcome the limitation of OLS in high dimensional setting, a regularization is done by adding an appropriate penalty over the regression coefficients. Most commonly used Lagrangian regularizations can be expressed in the form 
\begin{equation}
    \label{DSLw}
     (\hat{\boldsymbol{\beta}},\hat{\mathbf{\Delta}}_{1},...,\hat{\mathbf{\Delta}}_{g}) = \arg\min \frac{1}{2} \sum_{i} \left(y_{i}-\mathbf{x}_{i}^{T}(\boldsymbol{\beta} + \mathbf{\Delta}_{g_{i}})\right)^{2} + \lambda P(\boldsymbol{\beta},\mathbf{\Delta}_{1},\ldots,\mathbf{\Delta}_{g}), %\left(\|\beta\|_{1} + \sum_{g=1}^{G}r_{g}\|\mathbf{\Delta}_{g}\|_{1}\right)  ,
\end{equation}
 
\noindent where\\ $P(\boldsymbol{\beta},\mathbf{\Delta}_{1},\ldots,\mathbf{\Delta}_{g}) = \alpha_0 \sum_{g=1}^G\|\boldsymbol{\beta} + \mathbf{\Delta}_g\|_{0} + \alpha_1 \{\|\beta\|_1 + \sum_{g=1}^{G}r_{g}\|\mathbf{\Delta}_{g}\|_{1} \} + \alpha_2 \sum_{g=1}^G\|\boldsymbol{\beta} + \mathbf{\Delta}_g\|_{2}^2 $ (with $\|\beta\|_{q}^{q}$ = $\sum_{i=1}^{q}{|\beta_{i}|^{q}}$ for $q>0$ and $\|\beta\|_{0}$ $=$ $|\{i; \beta_{i}\neq 0 \}|$ ) for suitable $\alpha_i\geq0$ and user supplied adaptive weights $r_g$. The regularization parameter $ \lambda $ is usually chosen via some cross-validation scheme. Under $\ell_{0}$-penalty the estimator is infeasible to compute when $p$ is  large since the $\ell_{0}$-penalty is a non-convex function of $\boldsymbol{\beta}$ (\cite{sara}). Many other well known model selection criteria such as the Akaike Information Criterion (AIC), ridge regression (\cite{hoerl}), Lasso (\cite{tibshirani}), elastic net (\cite{zou}) and data enriched regression (\cite{chen}) fall into this framework. 

In \cite{adlasso} it has been observed that under $\ell_{1}$ penalty asymptotic setup is somewhat unfair, because it forces the coefficients to be equally penalized. Therefore it is advisable to use any other auxiliary information if we have one. In the IMDb example there is a polarity scores of each feature which turns out to be very useful auxiliary information. It turns out that the coefficients obtained through simple Lasso is highly correlated with polarity therefore we use inverse of polarity as a replacement of Lasso by adaptive Lasso in our algorithm. Therefore we can certainly assign different weights to different coefficients and if weights are data dependent and cleverly chosen, then adaptive Lasso can have the oracle properties. However the choice of the adaptive weights is an interesting problem which we demonstrate through the choice of six different weights and comparing them in our experimental results Section \ref{results}.

Using data shared AdaBag Lasso we study the grouped IMDb data in this article. We make use of the availability of polarity weights as obtained by \cite{potts} as an auxiliary information for the adaptive Lasso regularization.
\begin{equation}
 \begin{aligned}
     (\hat{\boldsymbol{\beta}},\hat{\mathbf{\Delta}}_1,\ldots,\hat{\mathbf{\Delta}}_g) = \arg\min \frac{1}{2} \sum_{i:g_i=g} \left(y_i - \mu-\mathbf{x}_{i}^T(\boldsymbol{\beta} +  \mathbf{\Delta}_{g_{i}})\right)^2 \\
     + \lambda \left(\|\mathbf{w}\cdot\boldsymbol{\beta}\|_1 + \sum_{g=1}^G r_g\|\mathbf{w}\cdot\mathbf{\Delta}_{g}\|_1\right)  ,
 \end{aligned}     
    \label{DSL1}
\end{equation}

where $\mathbf{w}=(w_1,w_2,\ldots,w_p)$ is the vector of \textit{inverse polarity weights}. Further given any two vectors $\mathbf{x}=(x_1,x_2,\ldots,x_p)$ and  $\mathbf{y}=(y_1,y_2,\ldots,y_p)$, $\mathbf{x}\cdot \mathbf{y}= (x_1y_1,x_2y_2,\ldots,x_py_p)$.
To retain numerical stability a small tolerance constant of the order of $10^{-5}$ has been added keep polarities away from zero.

Another popular dimension reduction method is principal component analysis (\cite{pcalda},\cite{jolliffe}). In a regression problem ($\mathbf{Y}=\mathbf{X}\boldsymbol{\beta}+\boldsymbol{\epsilon}$) using principal component procedure original predictor variables are transformed into new synthetic set of predictors (\textit{$v_1=\mathbf{u}_{1}^{T}\mathbf{x},v_2=\mathbf{u}_{2}^{T}\mathbf{x},\ldots$}) where $\mathbf{u}_i$'s eigen vectors of singular value decomposition of centered and scaled $\mathbf{X}$. Normally $\mathbf{u}_i$'s are ordered according to decreasing eigen values $\omega_1\geq\omega_2\geq\ldots$. The advantage of these new synthetic features is they are uncorrelated within the sample. A large number of these new set of predictors correspond to near-zero eigen values. Therefore to perform variable reduction by PCA some thresholding is necessary. We use '30\% variance explained' as thresholding limit in this article. We also introduced another variant of PCA where $\mathbf{u}_i$'s are ordered according to entropy of an artificial probability $(u_i^2,u_i^2,\ldots)$. This was done simply out of curiosity to check how much performance degrades in classification problems if we try to extract the \textit{most concentrated} PCs.

For classification problem the PC predictors are obtained from the training set $\mathcal{T}_r$. Based on the thresholded PCs we use the linear discrimination based on new synthetic set of predictors $\mathbf{v}=(v_1,v_2,\ldots,v_t)$ where $t$ is the number of PCs chosen for classification after thresholding. The final formula  $\phi(\mathcal{T}_r,\mathbf{x},g) = \phi(\mathcal{T}_r,\mathbf{v}^T\mathbf{b}) $ where $\mathbf{b}=\mathbf{S}_{\omega}^{-1}(\overline{\mathbf{v}}_1-\overline{\mathbf{v}}_2)$ which is the usual formula for Fisher's linear discrimination (\cite{jolliffe}).

The standard LDA can be seriously degraded if there are only a limited number of observations $n$ compared to the dimension of the feature space $p$ $(p>>n)$. To prevent this from happening it is recommended that the linear discriminant analysis be preceded by a principle component analysis. In PCA, the shape and location of the original data sets changes when transformed to a different space whereas LDA does not change the location but only tries to provide more class separability and draw a decision region between the given classes. Dimension reduction obtained through PCA lacks clarity in terms of tracking bag of words (\cite{pcalda}). Following a suggestion from the referee we use the role of PC-LD analysis to construct a baseline procedure for comparing with proposed data shared AdaBag Lasso for IMDb data.

\section{Methodology}

In this paper, the developed methodology produces the model with high polarity features that explain response variable better than \cite{gross} in terms of misclassification error and model size when $(p>>n)$. In addition, It reduces the dimension of the dataset which is achieved by considering minimum misclassification error at every step of the methodology. It conveys similar information concisely also takes care of multicollinearity, removes redundant features and fastens the time required for performing similar computations.

In AdaBag Lasso methodology we consider the estimation accuracy of the parameter $(\boldsymbol{\beta},\mathbf{\Delta}_g)$, a different task than prediction. Under compatibility assumptions on the design matrix $X$  in a linear model in G groups, let $\mathcal{S}_{g}$ denote the number of active set of variables
and the set of estimated variables using AdaBag as defined in \cite{sara} be given 
\begin{equation}
    \hat{\mathcal{S}}_g(\lambda) = \{j|\, (\hat{\boldsymbol{\beta}_j}+\hat{\mathbf{\Delta}}_{g,j})(\lambda)\neq 0 \,\,for\, j=1,\ldots,p,\,\lambda\in\Lambda \, \}.
\end{equation}
where $\Lambda $ is user supplied grid of $\lambda$.

In AdaBag lasso methodology the workflow is as follows
\subsection*{Principles of Methodology}
\begin{enumerate}
\label{adabag}
    \item
    \textbf{Data Organization.}
    Given the original sample $( y_i,\mathbf{x}_i^T ,g_i)$ where $\mathbf{x}_i \in \mathbb{R}^p$, $g_i \in \mathbb{G}= \{1,2,\cdots,G\}$,$i=1,2,\ldots,n$ which is a grouped dataset. We split it into training set consisting of $75\%$ of the observations and a test set consisting of $25\%$ of the observations. We transformed the data set into two classes. Both class observations are divided in $2:1:1$ ratio. The largest part goes core training set $\mathcal{T}_c$ and other parts goes to validation $\mathcal{V}_d$ and test $\mathcal{T}_s$ sets. Our training set consist of union of core training set and validation set. Initial model selection will be carried out in core set and optimization of the regularization parameter is done in the validation set.

    \item
    \textbf{Adaptive Lasso.}
    The sets of active set variables are generated using training set $\mathcal{T}_r$ after passing through AdaBag Lasso (\ref{DSL1}) over a grid of lambda of size $k$. Weights used in (\ref{DSL1}) are polarity scores (\cite{mass}, \cite{potts}) which returns $k$ nested sets $\hat{\mathcal{S}}_{1,g}(\lambda), \hat{\mathcal{S}}_{2,g}(\lambda),\ldots,\hat{\mathcal{S}}_{k,g}(\lambda)$ of high polarity features present in the dataset. We pool them together to consider the active set for future steps.
    
    \item
    \textbf{OLS Tuning.} 
    We tune the parameter obtained in step $2$ by running OLS on the set of active set variables $\hat{\mathcal{S}}_{k,g}(\lambda)$. The usefulness of this has already been discussed in \cite{olslasso}.

    \item
    \textbf{Prediction in $\mathcal{V}_d$ and modified Cross Validation.}
    The predicted responses $\{\hat{y_k}(\lambda),\ldots | k\in {\mathcal{V}_d}\}$ are computed using tuned coefficients $\left((\widehat{\boldsymbol{\beta}+\mathbf{\Delta}_g})^{OLS}(\hat{\mathcal{S}}_{k,g}(\lambda))\right)$ in step $2$ for different classes. 
    Predicted responses are segregated into two classes using classifier (\ref{mis_rule}). 
    \item
    \textbf{Optimization of Regularization Parameter $\lambda$.}
    We compute the misclassification error by varying $\lambda$ over $\Lambda$ where prediction in $\mathcal{V}_d$ using $\mathcal{T}_c$ generates average empirical misclassification error which is measured by $ME{(\lambda,\mathcal{V}_d)}=\frac{1}{|\mathcal{V}_d|}\, \sum_{l=1}^{|\mathcal{V}_d|} \, |y_l - \hat{y}_l(\lambda)| $. The  optimum value of $\lambda$ $(\hat{\lambda}_{opt})$ is chosen in step ($2$) by minimizing the misclassification error $ME{(\lambda,\mathcal{V}_d)}$ in validation set. 
    \textbf{This constitute one run of AdaBag Lasso when $\mathcal{T}_c$ is stratified}.
    
    \item
    \textbf{Bootstrap Resampling.}
    In order to create greater robustness and assign more relevance to different words chosen in word model we bootstrap the core training set($\mathcal{T}_c$) hundred times. We repeat step $2:5$ to obtain $\hat{\mathcal{S}}^{(*,1)}(\lambda^*_{opt,1})$,$\hat{\mathcal{S}}^{(*,2)}(\lambda^*_{opt,2})$,$\hat{\mathcal{S}}^{(*,100)}(\lambda^*_{opt,100})$ in step 2 by minimizing the empirical misclassification error attained in step 5.

    \item
    \textbf{AdaBag bfd.}
    Bagging frequency of each word is computed by $f^*(w)=(\sum_{l=1}^{100}\mathbf{\rm{I}}(w \in \hat{\mathcal{S}}^{(*,l)}(\lambda^*_{opt,l}))$, where $w \in bow$.
     Define the vector
    $$bfd = (f^*(w): w \in bow).$$
    \noindent This bfd provides a table where each word has empirical
    frequency obtained through steps $2:6$ (Fig.\ref{fig:bagofwords_comp}).
    We ordered the word according to the new empirical frequency. \textit{Next we choose different cutoff frequencies $c$ and vary it over 1:100 to get a collection of nested word models. This is the core output of the Data shared AdaBag Lasso methodology}. The output of AdaBag is given by
    $$\mathcal{A}_c = (w: f^*(w)\geq c)$$
    
    \item
    \textbf{Optimization over AdaBag sequence of models.} Optimize over cutoff $c$ in bfd and use OLS to obtain misclassification error over the same validation set $\mathcal{V}_d$ to get the minimum misclassification error in the validation set at cutoff level $c^* \in \{1,100\}$ (Fig. \ref{fig:cutoff_sim_group}).
    
    \item
    \textbf{Computation of Test Set Misclassification Error (TME).}
    Using the nested word model corresponding to $c^*$ that is word model $\mathcal{A}_{c^*} =( f^*(w)\geq c^* )$ is used to calculate the misclassification error in test set. Here we use shrinkage classifier for predicting in the test set.
\end{enumerate}
The above steps are expressed in algorithmic form below.
\begin{algorithm}{
%\Procedure{Steps}{}
%\State
\small

Read raw data\\
Split it into train set $\mathcal{T}_r(75\%) = (\mathcal{T}_c(50\%) \cup \mathcal{V}_d (25\%)),$ and $ \mathcal{T}_s (25\%)$ class wise in ratio 2:1:1.\\
 
\For{$z_{i}^* \in (z_{1}^*,z_{1}^*,\cdot\cdot\cdot,z_{100}^*)$}{
    Run Data Shared AdaBag Lasso over a grid of $k$ lambda values\\
     input: $(\lambda_{1}^*,\lambda_{2}^*,\cdots,\lambda_{r}^*)$\\
    \For{$\lambda_{k}^{*} \in (\lambda_{1}^{*},\lambda_{2}^{*},\cdots,\lambda_{k}^{*})$}{
    Run OLS on $\hat{S}_{k,g}(\lambda_g^*)$.\\
   Estimate $\hat{y_k}({\lambda_g}^*)$ in validation set $\mathcal{V}_d$.\\
   Segregate $\hat{y_k}({\lambda_g}^*)$ using classifier (\ref{mis_rule}).\\
   Calculate average prediction error $ME{(\lambda_g^*,\mathcal{V}_d)}=\frac{1}{|\mathcal{V}_d|}\, \sum_{j=1}^{|\mathcal{V}_d|} \, |y_j - \hat{y_j}(\lambda^*)| $.
    }
    Return $\hat{S}^{(*,i)}_g(\lambda_g^*)$ minimizing misclassification error $ME{(\lambda_{opt}^*,\mathcal{V}_d)}$ attained in $\mathcal{V}_d$.
}

 % $S^{**}_{AdaBag}=\left(\sum_{l=1}^{100}\mathbf{\rm{I}}(w \in \hat{\mathcal{S}}^{(*,l)}(\lambda^*_{opt})\right)$\\
  \For{$c \in1:100$}{
  Run OLS on word models $\mathcal{A}_c = \left(f^*{(w)}\geq c\right)$\\
  Estimate $\hat{y}^{OLS}_c$ in validation set $\mathcal{V}_d$.\\
   Segregate $\hat{y}^{OLS}_c$ using classifier (\ref{mis_rule}).\\
   Calculate average prediction error $ME({c,\mathcal{V}_d})=\frac{1}{|\mathcal{V}_d|}\, \sum_{k=1}^{|\mathcal{V}_d|} \, |y_k - \hat{y}_{k,c}({\lambda}^{OLS})| $.

  }
     Return $\mathcal{A}_{c^*} = \left(f^*{(w)}\geq c^*\right)$, where $c^*$ is minimizer of $ME({l,\mathcal{V}_d})$ in $\mathcal{V}_d$.

}
\caption{Data Shared AdaBag Lasso}
\label{euclid}
%\end{algorithmic}
\end{algorithm}
%\EndProcedure

\subsection*{Choice of Adaptive weights in Data Shared AdaBag Lasso Model}
\label{noise}
Choice of the regularization parameter $r_{g}$ in DSL model plays an important role in the sense that it controls the amount of sharing  within the groups. Here we propose new weighting schemes as function of sample sizes under the condition $\sum_{g}{r_{g}} > 1$. We tried different choices of the regularization parameter $r_g$ which are generated from the idea of De-noising proposed by \cite{donoho}.

Following \cite{donoho} suggested thresholding procedure for recovering functions for noisy data and applied soft thresholding nonlinearity $\eta_{y}(t) = sgn(y)(|y|-t)_{+}$ coordinate-wise to the empirical wavelet coefficient which is of the form
\begin{equation}
    \label{donoho}
     t_{n} = \sqrt{2log(n)}.\gamma_{1}.\sigma/\sqrt{n}  ,
\end{equation}
where $\gamma_{1}$ is a constant which is defined in \cite{donoho}.

The generalized data enriched model (\ref{dslopt}) can be written as
\begin{align*}
y_{i} &= \mu + x_{i}^{T}\mathbf{\beta} + x_{i}^{T}C_g\mathbf{\Delta}_g/C_g + \epsilon_{i}\\
 &= \mu + x_{i}^{T}\mathbf{\beta} + x_{i}^{T}\gamma_g/C_g + \epsilon_{i},
\end{align*}

where $\gamma_g$ = $C_g\mathbf{\Delta}_g$ 
are new group effects and $C_g$ is the variability order of the normalization column $X_{i}$.
The penalty term simplifies to
\begin{center}
$\lambda(C_{0}\|\boldsymbol{\beta}\|_{1} + \sum_{g=1}^{G}r_{g}C_{g}\mathbf{\Delta}_{g})$=
$\lambda(C_{0}\|\boldsymbol{\beta}\|_{1} + \sum_{g=1}^{G}m_{g}\mathbf{\Delta}_{g}$),
\end{center}
where $m_{g}$ = $r_{g}C_{g}$ are new weights over the groups. Standardization of columns should be done on different groups and pooled groups. Under normal thresholding when $\mathcal{O}(\sqrt{n}\hat{\boldsymbol{\beta}})$ are $\mathcal{O}_{P}(1)$ in such of case Donoho and Jhonstone find for declaring global threshold (\ref{donoho}) is of order $\sqrt{2log(n)}.\gamma_{1}.\sigma/\sqrt{n}$. Since we have sparse matrices so order may be little low of n and since order is not known, we tried different choice of n. From the following equation (\ref{donoho}) we have

\begin{center}
$\sqrt{\frac{n}{\log(n)}}=\sqrt{2}\gamma_{1}.\sigma/t_{n} \approx r_{g}$.
\end{center}

Different choice of $C_{0}$ gives different choice for the regularization parameter $r_{g}$. Following are the weighting schemes explored here\\
Let $N=\sum_{g=1}^{G}n_g$ then we have
\begin{itemize}
   \item 
    \textbf{WS 1}-Choosing $n\approx n_g \approx n$, $\log(n)\approx \log(N)$ and $C_{0} \approx N/\log(N)$ weights over groups becomes $\sqrt{1/3}$.
    \item
    \textbf{WS 2}-Choosing $n\approx n_g$, $\log(n)\approx \log(N)$ and $C_{0}$ $\approx$  $N/\log(N)$ weights over groups becomes $r_{g_i}$= $\sqrt{(n_g/N)}$.
    \item
    \textbf{WS 3}- By taking the square root of logarithm of square of the WS 1 leads to $r_g$= $\sqrt{\log{n_g}/\log{N}}$.
    
    \item
    \textbf{WS 4}- Squaring the WS 3 we get $r_g$= $\log{n_g}/\log{N}$.
   
    \item
    \textbf{WS 5}- By taking the reciprocal of WS 3 we get $r_g$= $\sqrt{\log{N}/\log{n_g}}$.
    
    \item
    \textbf{WS 6}- By dividing the WS 3 by WS 2 we get $r_g$= $\sqrt{\frac{\log{n_g}\times N}{\log{N}\times n_g}}$.

\end{itemize}
 Experimental results shows the empirical evidences for the choice of the regularization parameter which are discussed in detail in following Section \ref{results}. However we could not find any further theoretical justification for the choice of the adaptive weights and their performance which remains an interesting problem how to choose optimum weights. From our heuristics it turns out that $WS 3$ and $WS 4$ are performing uniformly superior to other schemes in terms of model size and misclassification error in test set ($\mathcal{T}_s$). We conjecture that these are close to theoretical mini-max bound for this problem. The heuristics through which these two weighting schemes have been derived that might lead to theoretical mini-max rate for this type of problem (as $n\rightarrow \infty , p \rightarrow \infty , p/n = \mathcal{O}(1)$ for sparse designed matrix $X$ with appropriate sparsity order).

\section{ Data Setup and Experimental Results}
\label{results}
We implemented our methodology on both IMDb review dataset and simulated data set and data analysis is performed for the various cases.
As there are $2^{p}$ possible sub-models, computational feasibility is crucial and when the data passes through the Lasso (\cite{tibshirani}) it selects at most n variables before it saturates in case of $(p>>n)$. To improve the computation feasibility of the model we need to compress the size of the bag of features.  The data shared AdaBag Lasso methodology can be implemented with any Lasso solver using a straightforward augmented data approach. The above problem is solved by R package \textbf{glmnet} (\cite{friedman}) which has the capability to use the sparse representation of a data matrix. It is quite faster than other Lasso solver. Here it is easy to change the family and loss function. In \textbf{glmnet} package we can fit ridge as well as elastic net by changing the values of $\alpha$. For ($\alpha = 1$) glmnet fits Lasso model, for $(\alpha = 0)$ it fits ridge regression and for $\alpha \in (0,1)$ it fits elastic net model.

\subsection*{Simulated Data}
For the simulation studies, we generate feature matrix $X$ with $1000$ observations and $77$ features from binomial family. Here columns of the feature matrix $X$ ($1:2,\,3:65,\,66:75,\,76:77$) are generated with binomial probability ($0.1,0.2,0.03,0.029$) respectively. 

To write the features matrix in the form $$y=\mu + \mathbf{x}\boldsymbol{\beta} +\epsilon$$ where $\epsilon$ is random error generated from Gaussian family with mean $0$ and variance ($0.3$). Now to generate response variable we assigned true $\beta$ weights vector $(-1, 1, -1, 1, 0.5, -0.5, -0.25,\\ -0.125, 1.25, -1.25)$ corresponding to features with indices $(1, 2, 3, 4, 5, 6, 7, 8, 76, 77)$ respectively, and other features are assigned weight $zero$.

We remove the observation which are lying between ($\bar{y}-\sqrt(0.3),\bar{y}+\sqrt{0.3}$) to divide the simulated data in two classes. Next, we split the data set into core training, validation and test set where ($200$ ,$100$, $100$)  observations are in training, validation and, test sets having equal number of observations from both class.

To implement AdaBag Lasso on simulated data polarity scores to features are assigned in a sequence ($1:8,9:21,22:25,26:75,76:77$) with scores ($a,b,c,d,e$) respectively. Here $a=(-1, 1, -1, 1, 0.5, -0.5, -0.25, -0.125$), $b$= ($0.25, 0.33, 0.33, 0.20, 0.20, 0.25, 0.20, 0.20, 0.20,0.20, \\0.20, 0.20$), $c=(0.2,0.2,0.2,0.2)$, $d=(\frac{1}{i}:i \in (25,26,\ldots,75))$, and $e = (1.25,1.25)$. In AdaBag Lasso we use weights as reciprocal of absolute value of polarity scores. To retain numerical stability, a small tolerance constant of the order of $10^{-5}$ has been added keep polarities away from zero. The experimental results are discussed in detail below.

\begin{figure}[htbp!]
\centering
\includegraphics[width=150mm,height=25mm]{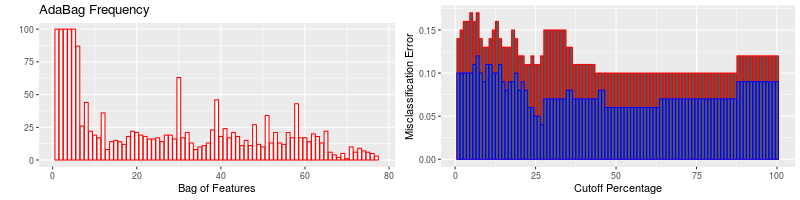}
\caption{The left panel is plot for bagging frequency in hundred bootstrap samples assigning equal polarity to each feature. The right panel shows the misclassification error attained with varying cutoffs in validation and test set in simulated data.}
\label{fig:cutoff_sim_lasso}
\end{figure}

We performed data analysis using AdaBag Lasso methodology on simulated data with above setup on two variations of polarity scores. Firstly we analyze the dataset by assigning equal polarity scores to features and in second variation we analyze by randomly assigning polarity scores to each features. Right panel in Fig. \ref{fig:cutoff_sim_lasso} is showing the bagging frequency of each feature and right panel shows the validation and test set misclassification error at different cutoff percentage. Red columns in right panel of Fig.\ref{fig:cutoff_sim_lasso} shows the range of misclassification error in test set and blue columns shows the range of misclassification error in validation set. In the right panel of Fig.\ref{fig:cutoff_sim_lasso} it seems that global minima cutoff attained in validation set ($4\%$ ME) returns a slightly higher misclassification error in test set($12\%$) with features ($\textbf{1,  2 , 3,  4 , 5 , 6 , 8 },12, 30, 39, 48, 51, 58$). Here it captures $7$ true features and  $6$ redundant features. However a second minima attained in validation set($6\%$) towards higher frequency features returns a global minima in test set($10\%$) with features ($\textbf{1 , 2 , 3,  4,  5,  6 },30$). Here it captures $6$ true features  and $1$ redundant feature in the model. From both the set of feature model it seems that choice of cutoff in bagging frequency needs to be optimized.

\begin{figure}[htbp!]
\centering
\includegraphics[width=150mm,height=25mm]{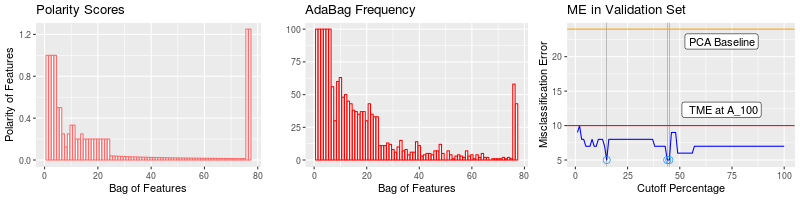}
\caption{The left panel is plot for polarity assigned to each feature. Middle panel is plot for bagging frequency in hundred bootstrap samples. The right panel shows the misclassification error attained with varying cutoffs in validation set in simulated data.}
\label{fig:cutoff_sim}
\end{figure}

Left panel in Fig. \ref{fig:cutoff_sim}, is showing the polarity scores assigned to each feature, middle panel shows the bagging frequency of each feature, and right panel shows the misclassification error attained in validation set at different cutoffs $c$. From the first two panel we see that high polarity features attain higher frequency in bfd.

In the right panel of Fig \ref{fig:cutoff_sim} we see that minimum misclassification error (of $5\%$ level) in validation set is attained at three different cutoffs $c_1,c_2,c_3$. First cutoff $c_1$ captures following features from bfd $(1 , 2  ,3,  4,  5,  6 , 7 , 8 , 9, 10, 11,12, 13 ,14, 15, 16, 17, 18, 19 ,20 ,21, 22, 23, 24, 33, 76, \\77)$ which returns $13\%$ misclassification error in test set while second and third cutoffs capture same features from bfd $(1,  2,  3,  4,  5,  6,  7,  9, 10, 11, 12, 13, 76)$ which returns $12\%$ misclassification error in test set. 

In such case, we note that, given the core set the number of observation are more than features therefore it is possible that we have not observed the effect of curse of dimensionality. The second minima is attained at cutoff level $45$, although it is missing few true beta features but still is the smallest model and giving smallest misclassification error. In validation set we have got one more minimizer which are going down upto $5\%$ error but corresponding models have more redundant variables.

In right panel of Fig. \ref{fig:cutoff_sim} two horizontal lines are included. The top horizontal line indicate the misclassification error attained at variance ordered PCs which we use as baseline to compare proposed methodology. The bottom line indicate the test set misclassification error attained at $\mathcal{A}_{100}$. However AdaBag Lasso captures top $6$ true features ($1,2,3,4,5,6$) with bagging frequency $\mathcal{A}_{100}$ and returns $10\%$ misclassification error in test set. 

To compare the efficiency of the AdaBag Lasso method we performed PCA and LDA on simulated data. Here simulation results shows that to explain $30\%$ of total variation we need $11$ variance ordered PCs and $16$ entropy ordered PCs with $32\%$ and $37\%$ error rate in test set. From above Fig. \ref{fig:cutoff_sim} and test set results we see that AdaBag Lasso performs better the PCA and LDA in terms of misclassification error.

\subsection*{IMDb Data }
The IMDb review dataset aclImdb \cite{mass} of movie reviews from ~\url{IMDb.com} contains set of reviews and corresponding ratings associated with them. Here the dependent variable $Y_{i}$'s are integer rating and explanatory variable $X_{i}$'s are text features. The text features are converted into integers by taking the number of occurrences of each feature in each review. This resulted in a high dimensional dataset that is the sample size is less than the features. Later the entries of the predictor matrix $X$ are converted into the binary matrix and the sparse representation of the data matrix $X$ is used for the analysis.
\begin{equation*}
  x_{ij}=
    \begin{cases}
    1, & \text{if $j^{th}$ feature is present in $i^{th}$ review }\\
    0, & \text{otherwise}.
    \end{cases}
\end{equation*}

This dataset contains 50k polarized movie reviews where half of the reviews are positive (rating $\geq7$) and other half are negative (rating $\leq4$), allowing no more than 30 reviews per movie. The proposed model fitted over three most popular genres drama, comedy, and horror. Only those review are considered which appear in at least one of the three genres. Only those words that appear in at least five reviews in the dataset are considered for the analysis of the dataset which resulted in $p = 27743$ features in our dataset. To implement the proposed methodology, we mainly focused on genres of the movies.

To perform the analysis the whole dataset is divided into three parts training, validation and test sets where reviews of both the classes are distributed in the ratio 2/1/1 respectively. This results in 8561, 4281,4281 observations in Drama, 4901,2449,2451 observations in comedy and 3235,1619,1617 observations in horror, in train, validation and test sets respectively.

\begin{figure}[htbp!]
\centering
\includegraphics[width=150mm,height=40mm]{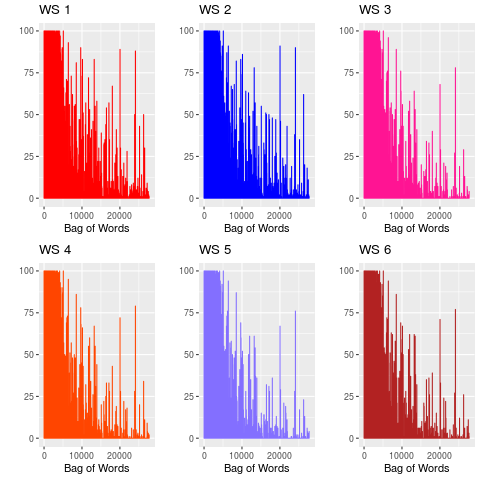}
\caption{Bagging frequency of features after running AdaBag Lasso methodology on IMDb data using different weighting schemes in DSL. Here x-axis represents the bag of features and y-axis represents the bagging frequency of features.}
\label{fig:bagofwords_comp}
\end{figure}

Next, we implemented AdaBag Lasso on IMDb dataset using different weighting schemes explored in \ref{noise} in DSL. Fig. \ref{fig:bagofwords_comp}, shows the \textit{bfd} of words in hundred bootstrap samples. Here we see that the distribution of features are throughout the dataset and \textbf{ one-time Lasso run fails to capture these features. The AdaBag Lasso method helps us to find the relevance of features in high dimensional variable selection problems and allows to choose features which take parts in explaining response variable in most of the bootstrap samples}.
 
\begin{figure}
\centering
\includegraphics[width=120mm,height=90mm]{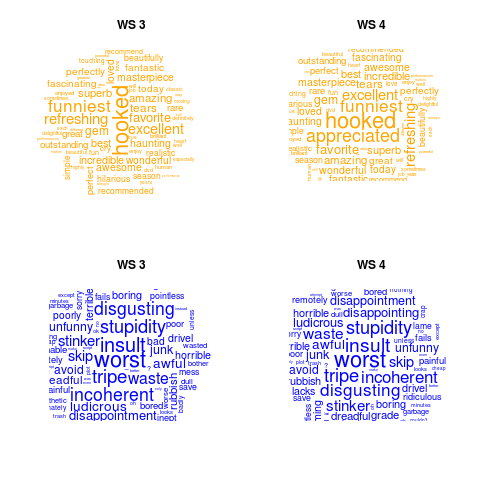}

\caption{Word cloud for top two best performing weighting schemes in 
 data shared AdaBag Lasso model with bagging frequency hundred. The orange color indicates the wordcloud for the positive sentiments and blue color indicates the negative sentiments based on OLS coefficients to top words taken from bfd.}
\label{fig:word_cloud}
\end{figure}

\begin{figure}[htbp!]
\centering
\includegraphics[width=140mm,height=40mm]{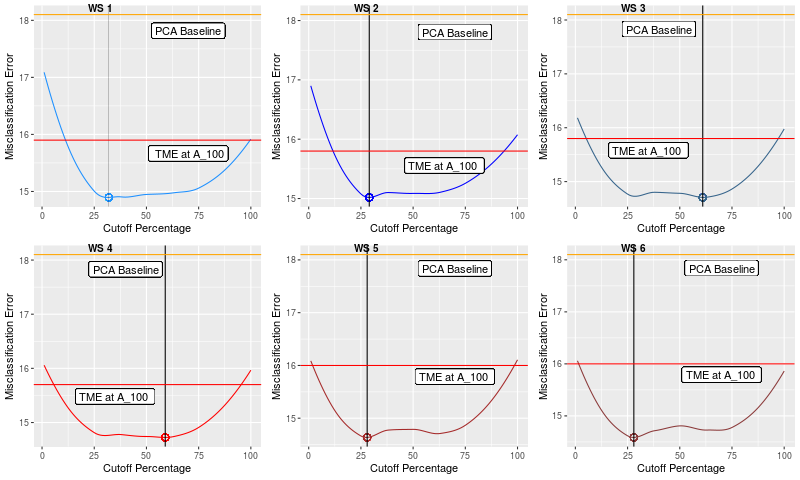}
\caption{Misclassification error attained at different cutoffs in bagging frequency for different weighing schemes in DSL. Here x-axis represents the cutoff percentage and y axis represents the misclassification error attained at different cutoffs.}
\label{fig:cutoff_sim_group}
\end{figure}

In Fig.\ref{fig:cutoff_sim_group} shoes a loess (\cite{cleveland}) regression fit to misclassification error in validation set for all weighting schemes discussed in \ref{noise}. Here we see that there exist a unique minimum for all the weighting schemes in validation set (vertical line added corresponding to global minima), which allow us to choose the optimum word model from \textit{bow}. Test set results for the optimum choice of cutoff $c^*$ are explored Table \ref{table:332}. From the Table \ref{table:332} we see that weighting schemes (4 \& 5 ) perform better in terms of TSME and model size.  

Two horizontal lines are added in Fig \ref{fig:cutoff_sim_group}. Top horizontal line indicate the PCA baseline and bottom horizontal line indicate the test set misclassification error attained at word model $\mathcal{A}_{100}$. From both the horizontal lines we see that AdaBag performs better than PCA and LDA algorithms and also TSME at $\mathcal{A}_{100}$ is close to validation set misclassification error at $\mathcal{A}_{100}$. 

Table \ref{table:2w} explores the performance of PCA and LDA algorithms on IMDb dataset. Here we run LDA algorithm on both variance ordered PCs and entropy ordered PCs. Test set misclassification error along-with number of PCs required to explain significant amount of variance in variance ordered PCs and entropy ordered PCs are given in Table \ref{table:2w}. Here we see that to attain $18.1\% , 25.7\%$ ME in pooled data we need $161$ variance ordered PCs and $ 219$ entropy ordered PCs which explains the $30\%$ of the total variations. However proposed AdaBag Lasso model performs better in terms of misclassification error and running PCA algorithm on high dimensional data is time-consuming too.

\begin{table}[htbp]
\begin{minipage}{1\textwidth}
\centering
{\small
\begin{tabular}{|ccccccc|}
    \hline
   WS&Model&Pooled&Drama&Comedy&Horror &Model Size \\
    \hline
   
    1 &AdaBag &  13.4 & 12.7 &   13.3  & 15.2 &  724 \\
   %  \hline\hline
   
    2 &AdaBag &  13.3   &    12.6&     13.1 &  15.2  &  833 \\
   %\hline\hline
    3&AdaBag& 12.8 & 12.2 & 12.7 & 14.5 &   447 \\
   %   \hline\hline
    4&AdaBag& 13.1   &   12.4  &     13.3  &  14.9   &  432 \\
   %\hline\hline
    5  &AdaBag & 13.2 & 12.9 & 12.6 & 14.9 &  887 \\
  %  \hline\hline
    6&AdaBag& 13.2  &    13.1   &   12.6 &  14.8  &  828\\
   \hline

\end{tabular} 
}
\end{minipage}
\vskip10pt
\caption{Test set misclassification error at different cutoffs for different weighting schemes after running AdaBag Lasso model.}
\label{table:332}
\end{table} 

\begin{table}[htbp]
\begin{minipage}{1\textwidth}
\centering
{\small
\begin{tabular}{|cccccc|}
    \hline
     &&Pooled&Drama&Comedy&Horror  \\
     %\hline\hline
     Prop. of Var Exp.&Model&ME(NOPC)&ME(NOPC)&ME(NOPC)&ME(NOPC)  \\
     \hline
      20  &Variance PC&  19.8(79) & 20.3(75) & 20.6(65)&20.9(65) \\
    % \hline\hline
     25 &Variance PC&  19.4(119) & 18.4(110) & 19.2(90)&20.7(90) \\
    % \hline\hline
     30 &Variance PC&  18.1(161) & 16.9(150) & 19.5(125)&20.0(120) \\
    % \hline\hline
     20  &Entropy PC&  38.4(120) & 39.3(117) & 25.8(94)&18.9(92) \\
    % \hline\hline
     25 &Entropy PC&  37.2(165) & 25.9(157) & 18.2(124)&18.5(126) \\
    % \hline\hline
     30 & Entropy PC& 25.7(219) & 17.3(200) & 16.9(166)&19.2(136) \\
     \hline
\end{tabular} 
}
\end{minipage}
\vskip10pt
\caption{Test set misclassification error using Principal components with Linear Discrimination for usual variance ordered and entropy ordered on IMDb data set.}
\label{table:2w}
\end{table} 

\begin{figure}
    \centering
    \includegraphics[width=13cm,height=8cm]{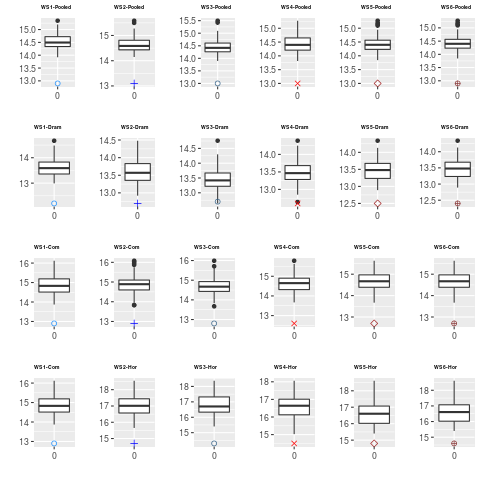}
    \caption{Box plot of test set misclassification error on each bootstrap optimized active set of variables using validation set for different weighing schemes. The point in each box plot indicates the test set misclassification error attained at bagging cutoff $c*$.}
    \label{boxplot}
\end{figure}

In addition the distribution of misclassification error in test set using bootstrap optimized set of active set of variables (set of active set of variables optimized using validation set) are graphically presented in Fig. \ref{boxplot} for different weighting schemes. Also, TSME at optimum cutoff $c^*$ is presented in in each panel of the plot. Here we see that word models chosen from bagging frequency produces significantly improved classification error rate in test set.

Finally to demonstrate the relevance of the AdaBag procedure we provide the word cloud which shows that smallest AdaBag frequency $\mathcal{A}_{100}$ Fig.\ref{fig:word_cloud}. Word cloud (\ref{fig:word_cloud}) is generated using words models $\mathcal{A}_{100}$ for two top performing weighting schemes (WS 3 \& WS 4) in test set. Top hundred features taken from bfd and are plotted in order to OLS coefficient where size of each word represents the absolute value of the coefficient. Positives and negative sentiment words are plotted in order to sign of coefficient is indicated by color blue and orange respectively.

\section{Concluding Remarks}

In this paper, we introduce Data Shared AdaBag Lasso methodology to reduce the dimension of the data set. This was demonstrated by a simulated data and a real word sentiment analysis problem.

Our simulation results has shown that by assigning equal polarity score to each feature results alike misclassification error in test and validation set at different cutoffs. Right panel of Fig. \ref{fig:cutoff_sim_lasso} shows that global minima attained in both sets contain almost same feature model in bfd. However Fig. \ref{fig:cutoff_sim} shows that if randomly assign some polarity scores to each feature then it is observed that high polarity features are having higher frequency in bagging table.

From the above applications explored in \ref{results}, it seem that the choice of second minima towards high frequency models returns a lower TSME and also a much lower dimension model. However optimum choice of cutoff $c^*$ still has to be optimized. Since we propose to choose the global minimum that's why we did not study the results for each weighting schemes in both the local minima. It is quite possible that the existence of two minima are pointing towards a nonlinearity where a smaller model explains the major part of the data. However a second linear combination consisting of more words might refine it and further reduce the misclassification error in test set. However we have not studied such nonlinear rules involving multiple linear classifier in this article. In this article we consider only one linear classifier but important linear classifiers are emerging from our figures and tables, particularly Fig. \ref{fig:cutoff_sim_group} which indicates important low dimensional linear combinations. Possible a nonlinear procedure involving multiple linear classifiers would reduce the misclassification error further.

Choice of adaptive regularization parameter $r_g$ can be made on the basis of minimum TSME and model size. However we have not provided any theoretical justification for choice of regularization parameter. The idea modified cross validation significantly improves the results in terms of model size and TSME.

It is natural to ask data shared AdaBag Lasso compares to other dimension reduction techniques. The most popular of these is principal components analysis (PCA). PCA finds new dimensions that explain most of the variance in the data. It is best at positioning those points that are far apart from each other because they are the drivers of the variance. It is also a linear method, meaning that if the relationship between the variables is nonlinear it performs poorly.Creation of word cloud using PCA is also a difficult task. While our methodology uses filtration at each step and high relevance features with higher polarity are retained in the model.

\section*{Acknowledgement}
The author would like to thank Professor Debapriya Sengupta, Applied Statistics Unit, Indian Statistical Institute, Kolkata, for his constructive collaboration in this work.

\end{document}